\documentclass[conference]{IEEEtran}
\IEEEoverridecommandlockouts
\usepackage{cite}
\usepackage{amsmath,amssymb,amsfonts}
\usepackage{algorithmic}
\usepackage{multirow}
\usepackage{dblfloatfix}
\usepackage{graphicx}
\usepackage{textcomp}
\usepackage{xcolor}
\usepackage{numprint}
\usepackage[bookmarks=false]{hyperref}
\def\BibTeX{{\rm B\kern-.05em{\sc i\kern-.025em b}\kern-.08em
    T\kern-.1667em\lower.7ex\hbox{E}\kern-.125emX}}
\npdecimalsign{\ensuremath{.}}

\DeclareMathOperator*{\vmax}{\max\vphantom{i}}

\begin{document}

\title{Deep Metric Learning using Similarities from Nonlinear Rank Approximations}



\author{
Konstantin Schall, Kai Uwe Barthel, Nico Hezel, and Klaus Jung\\
\textit{Visual Computing Group, HTW Berlin}\\
Berlin, Germany\\ 
{\{schallk, barthel, hezel, jungk\}@htw-berlin.de}
}


%
\maketitle

%
\begin{abstract}
In recent years, deep metric learning has achieved promising results in learning high dimensional semantic feature embeddings where the spatial relationships of the feature vectors match the visual similarities of the images. Similarity search for images is performed by determining the vectors with the smallest distances to a query vector. However, high retrieval quality does not depend on the actual distances of the feature vectors, but rather on the ranking order of the feature vectors from similar images. In this paper, we introduce a metric learning algorithm that focuses on identifying and modifying those feature vectors that most strongly affect the retrieval quality. We compute normalized approximated ranks and convert them to similarities by applying a nonlinear transfer function. These similarities are used in a newly proposed loss function that better contracts similar and disperses dissimilar samples. Experiments demonstrate significant improvement over existing deep feature embedding methods on the CUB-200-2011, Cars196, and Stanford Online Products data sets for all embedding sizes.
\end{abstract}

%

\section{Introduction}
Deep neural networks have led to significant improvements in a wide range of visual computing tasks such as image classification, segmentation and retrieval \cite{Russakovsky14, He15, Babenko14}. In traditional classification networks, the original input pixels are transformed into semantic and robust representations, describing the characteristic of an input image in order to distinguish it from different classes. However, real world visual data is very diverse and it is often not sufficient to describe detailed concepts with high level class labels. 

The research field of deep metric learning tries to find a representation space, in which the visual similarity of images is encoded by the distances of respective feature vectors produced by a neural network \cite{Hoffer14, Song16, Sohn16, Wang17}. Simple nearest neighbor search can then be used for fine grained classification tasks as face recognition \cite{Schroff15, Lu17}. Content based image retrieval is another field that greatly benefits from metric learning. High retrieval quality requires visually similar images (from the same class) to be mapped to similar feature vectors. Deep metric learning based loss functions therefore directly try to optimize the resulting embedding space by enforcing the distances between samples of the same class to be smaller than distances between entities of different classes. 

A number of loss functions have been proposed to achieve this goal \cite{Hadsell06, Hoffer14, Song16, Sohn16, Wang17}. The main improvements over the original ideas of Contrastive loss \cite{Hadsell06} and Triplet loss \cite{Schultz03} have been accomplished by evaluating vector constellations with respect to the entire batch instead of pairs and triplets. This allows a more global view of the actual embedding space and helps to solve difficult local constellations, leading to improved retrieval and clustering results. All these loss functions directly work with the distances in the embedded feature vector space.
However, nearest neighbor based problems, as image clustering and retrieval are evaluated by a ranking order. It is therefore not important to have very small distances to all samples of the same class, as long as all these distances are smaller than those of other classes. 
Instead of calculating the loss from the actual distances, we map all distances to normalized approximated ranks in the interval $[0, 1]$. These rank approximations are then converted to similarities using a nonlinear transfer function, where lower ranks lead to higher similarities. The final loss function is calculated for all batch samples with respect to their two most problematic samples of the batch: the sample from the same class with the lowest similarity and the sample from another class with the highest similarity. These steps allow to optimize each batch from a nearest neighbor perspective, which leads to well distributed embeddings. Our proposed \textit{Nonlinear Rank Approximation (NRA)} loss shows significant improvements compared to other loss functions used for deep metric learning and achieves new state-of-the-art results on the CUB-200-2011 \cite{CUB_200_2011} data set. For the Cars196 \cite{Car196} and Stanford Online Products \cite{Song16} data sets our approach is only outperformed by an attention based ensemble approach \cite{Kim18} or by more powerful neural networks as used in \cite{Manmatha17}. \IEEEpubidadjcol

\section{Related Work}
There are a number of other approaches, like Boosting of Independent Embeddings (A-BIER) \cite{Opitz18}, Hard Aware Deeply Cascaded Embedding (HDC) \cite{Yuan17}, Attention Based Ensemble (ABE) learning \cite{Kim18} or Query Expansion \cite{Azad17}, which can be combined with any loss function to further improve the image retrieval quality. However, in this work we focus on improving deep metric learning based loss and mining functions.

Let $\mathcal{X}$ be the set of input data with samples $\boldsymbol{x} \in \mathcal{X}$. A neural network with parameters $\boldsymbol{\theta}$ takes $\boldsymbol{x}$ and generates the embedding $f(\cdot; \boldsymbol{\theta}) : \mathcal{X} \rightarrow \mathbb{R}^d$. 
We omit $\boldsymbol{\theta}$ for simplicity and call $f(\boldsymbol{x})$ the feature vector of $\boldsymbol{x}$. The $L^2$ distance of two feature vectors $f(\boldsymbol{x}_i)$ and $f(\boldsymbol{x}_j)$ is denoted by $D_{i,j} = \| f(\boldsymbol{x}_i) - f(\boldsymbol{x}_j) \|_2$.

\textbf{Contrastive Loss} \cite{Hadsell06} takes two samples $f(\boldsymbol{x}_i)$ and  $f(\boldsymbol{x}_j)$ and tries to minimize their $L^2$ distance if they have the same class label and to maximize otherwise \cite{Hadsell06}. 
The loss function is formulated in (\ref{eq:ContrastiveLoss}), where $\tilde{y}_{i,j}$ is 1 if the images are from the same class and 0 otherwise.
\begin{equation}
\label{eq:ContrastiveLoss}
J_{i,j} = \tilde{y}_{i,j} D_{i,j}^2 + (1 - \tilde{y}_{i,j})\max{(0, \alpha - D_{i,j})}^2
\end{equation}
The optimal solution of such embeddings is reached, if all distances to positive images are zero and the negative distances are greater than a given margin parameter $\alpha$. This is very hard to achieve and a loss formulation based on actual distances leads to bad convergence properties.

\textbf{Triplet Loss} \cite{Schultz03} builds a set of three samples $\mathcal{T} = \{\boldsymbol{x}^a, \boldsymbol{x}^+, \boldsymbol{x}^-\}$ which consists of an anchor  $\boldsymbol{x}^a$, one positive example $\boldsymbol{x}^+$ of the same class as the anchor and one negative example  $\boldsymbol{x}^-$ with a different class. The loss value for a single triplet is calculated by
\begin{equation}
\label{eq:TripletLoss}
J_{a, +, -} = \max{(0,D_{a,+}^2 - D_{a,-}^2 + \alpha})
\end{equation}
This method generally performs better than Contrastive Loss, since distances are not directly minimized but rather the ratio of the negative and the positive distance w.r.t the anchor feature vector $f(\boldsymbol{x}^a)$ is maximized here.

\textbf{Lifted Structured Loss} \cite{Song16} was the first method to introduce \textit{importance sampling} for building batches. However, each batch with $m$ samples focuses on one single class only. $k$ positive samples are chosen and the remaining $m - k$ elements are hard (difficult) negative examples. All negative batch elements $\mathcal{N}$ are weighted by their distances to each positive entity $\mathcal{P}$ with the exponential function and the final loss value per batch is calculated as 
\begin{align}
\tilde{J}_{i,j} &= \log\Big(\!\sum_{(i,k) \in \mathcal{N} } \!\!\!\!\exp( \alpha - D_{i,k})+\!\!\!\!\!\sum_{(j,l) \in \mathcal{N} } \!\!\!\!\exp( \alpha - D_{j,l})\Big) + D_{i,j} \nonumber\\
J &= \frac{1}{2|\mathcal{P}|}\sum_{(i,j) \in \mathcal{P} } \max(0, \tilde{J}_{i,j})^2
\label{eq:LSLoss}
\end{align}
Focusing on a single or a small number of classes and their neighbors optimizes only a local region of the entire embedding space and can lead to suboptimal constellations.

\textbf{N-pair Loss}: To overcome this problem, \cite{Sohn16} introduced a multi-class N-pair loss function that uses $N$ pairs of $N$ different classes, where $N$ is desired to be as close as possible to the total number of classes.
This sampling ensures each data point to have exactly one positive example and $N-2$ negative ones. Similar to the Lifted Structured approach, the entire batch can then be used to calculate a loss value,
but instead of the $L^2$ distance the dot product is used as similarity measure:
\begin{equation}
\label{eq:NPairLoss}
J = \frac{1}{N}\sum_{i=1}^N \log\!\Big(\!1+\!\sum_{j \ne i} \exp(f(x_i)^T\!f(x_j) - f(x_i)^T\!f(x_i^+)\!\Big)
\end{equation}
Ignoring other positive data points and calculating the loss with respect to one single positive distance will always result in a low loss value, if the positive distance is already small in the N-pair and Triplet approaches. For N-pair one batch with a large positive distance results in a much higher loss value and leads to a more informative gradient, yet with the sampling of pairs, such constellations are only found occasionally.  
\begin{figure*}
  \includegraphics[width=\linewidth]{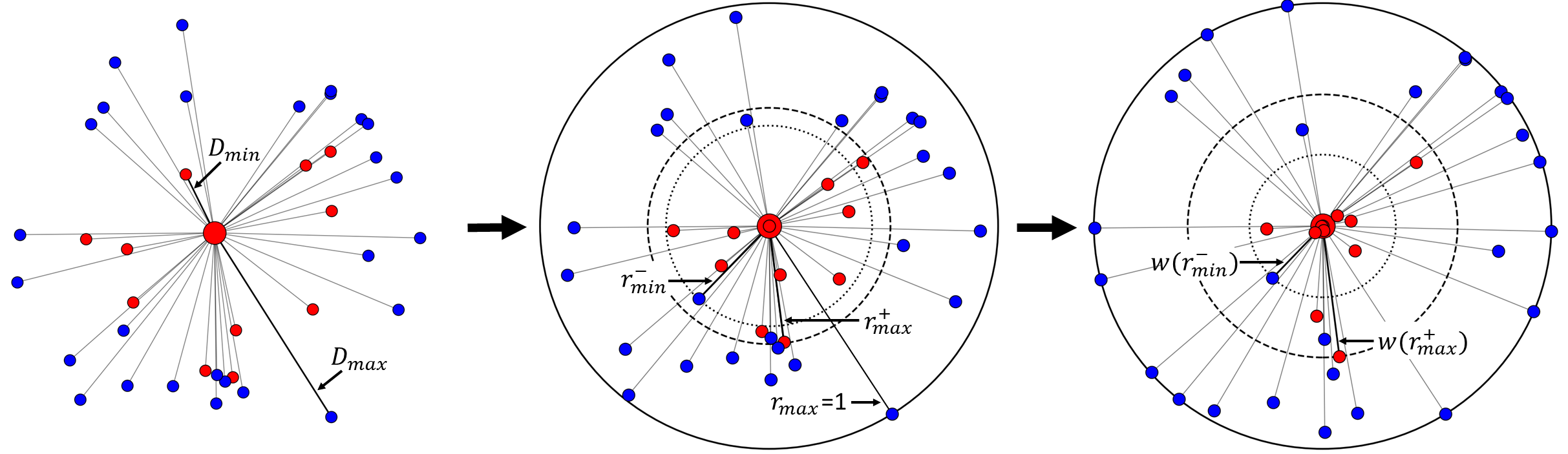}
  {\footnotesize\begin{tabular*}{\linewidth}{@{\hspace{4em}}c@{\hspace{11.7em}}c@{\hspace{6.2em}}c@{\extracolsep{\fill}}}
    (a) Anchor distances & (b) Normalized approximated ranks & (c) Nonlinear transformed normalized ranks
  \end{tabular*}}
  \caption{(a) Example embedding for an anchor point (shown as larger dot) from one class (red) and points from other classes (blue). For the sake of clarity, the indices $i,j$ are omitted. (b) All distances  $D$ are converted to normalized approximated ranks $r \in [0, 1]$. (c) These ranks are nonlinearly transformed $w(r)$ to  determine the similarities $s = 1 - w(r)$. Perfect clustering for an anchor point is only achieved if all distances of points from other classes are greater than the maximum distance of all points of the same class. Consequently our loss function uses the two most important similarities 
  $s^{+}_{\vmax}$ and $s^{-}_{\min}$. }
  \label{fig:teaser}
\end{figure*}
\section{Proposed Method}
\begin{figure*}
    \centering
    \includegraphics[width=\linewidth]{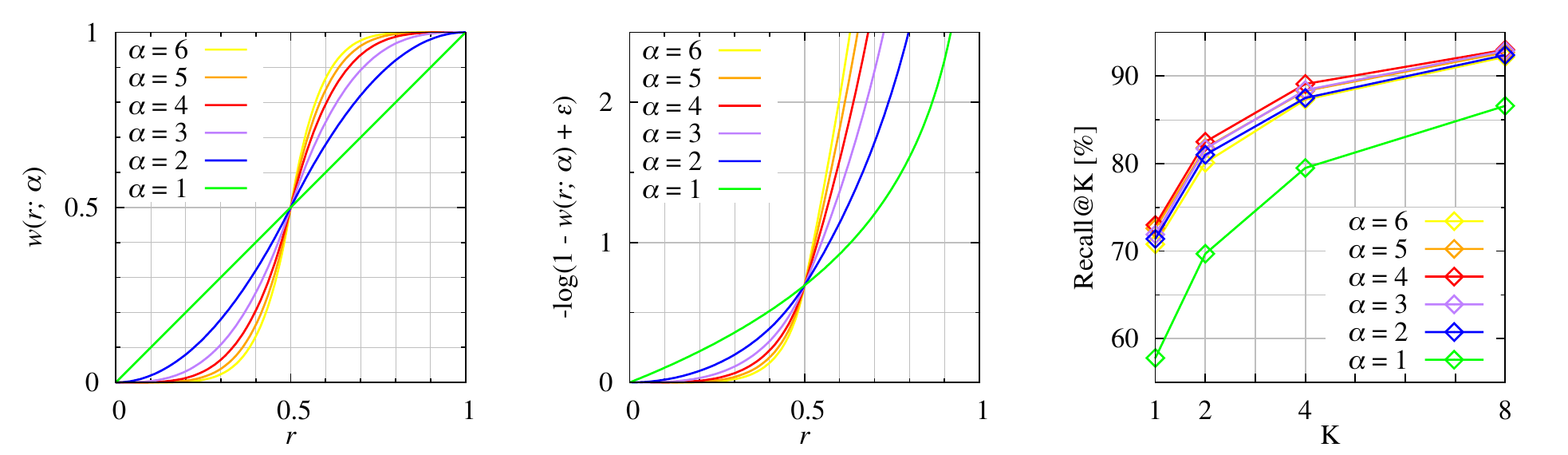}
    \caption{Transfer function (left), the corresponding loss component (center), and Recall@K results on the Cars196 data set (right)}
    \label{fig:transfer-function}
\end{figure*}

Opposed to the described approaches we build batches that contain a larger number of positive examples. This allows finding samples of the same class that are poorly embedded more easily. We use $m = k \cdot n$ samples $\boldsymbol{x}_i \in \mathcal{X}, \, i \in \{1, 2, \ldots, m\}$ of the input data per batch, where $k$ is the number of classes and $n$ the number of samples per class. 
Subsequently, we refer to this technique as \textit{uniform n-group sampling}. In rare cases where a class does not have enough representative images, more classes are added until the desired batch size is reached.
The set of positive samples $\mathcal{P}_i$ and the set of negative samples $\mathcal{N}_i$ w.r.t. an anchor point $\boldsymbol{x}_i$ with class $y_i$ 
are given by
\begin{equation}
\mathcal{P}_i = \{j \, | \, j \ne i \land y_j = y_i \} \quad \mathcal{N}_i = \{j \, | \, y_j \ne y_i \}
\end{equation}
$
\mathcal{S}_i = \mathcal{P}_i \cup \mathcal{N}_i
$ 
is the set of all batch samples excluding the anchor.
Instead of using a limited set of anchors, all samples of the batch serve as anchors. For each anchor's feature vector $f(\boldsymbol{x}_i)$ the distances $D_{i,j}$ to all other feature vectors of the batch elements are determined. $D_{i,\max}$ is the maximum of these distances, whereas $D_{i,\min}$ is the minimum distance to all other feature vectors of the batch w.r.t. this anchor:
\begin{equation}
D_{i,\max} = \max_{j \in \mathcal{S}_i } D_{i,j} \quad  D_{i,\min} = \min_{j \in \mathcal{S}_i }  D_{i,j} 
\end{equation} 
As high retrieval quality does not depend on the actual distances and positions of the feature vectors in the embedded space, but rather on the preserved ranking order, we focus on identifying those feature vectors that most strongly affect the retrieval quality. 
Following the assumption that a perfect clustering is achieved if and only if all distances to negative examples are larger than the maximum distance to positive examples, we search for the two most important distances w.r.t. an anchor: The maximum distance to the embeddings of its positive examples and the minimum distance to its negative examples.  
\begin{equation}
D^+_{i,\max} = \max_{j \in \mathcal{P}_i } D_{i,j} \quad D^-_{i,\min} = \min_{j \in \mathcal{N}_i } D_{i,j}
\end{equation}
Instead of using these distances directly, we use approximated normalized ranks. For each anchor $f(\boldsymbol{x}_i)$ the normalized ranks $r_{i,j}$ for the remaining batch elements are approximated by 
\begin{equation}
r_{i,j} = \frac{D_{i,j} - D_{i,\min}}{D_{i,\max} - D_{i,\min}} \in [0, 1]
\label{eq:normalized-rank}
\end{equation}
Accordingly we determine the approximated normalized ranks for the most distant embedding of the same class $r^+_{i,\max}$ and the closest embedding of a sample of a different class $r^-_{i,\min}$
\begin{equation}
r^+_{i,\max} = \frac{D^+_{i,\max} - D_{i,\min}}{D_{i,\max} - D_{i,\min}}, \:\: r^-_{i,\min} = \frac{D^-_{i,\min} - D_{i,\min}}{D_{i,\max} - D_{i,\min}}
\end{equation}
Next for each anchor we transform these ranks to similarities. Instead of calculating the similarity as $s = 1 - r$ we introduce a parametric nonlinear version using a transfer function
\begin{align}
w(r; \alpha) &= 
\begin{cases}
\frac{1}{2}(2r)^\alpha & r \in [0, \frac{1}{2}) \\
1 - \frac{1}{2}(2(1-r))^\alpha & r \in [\frac{1}{2}, 1] \\
\end{cases}
\label{eq:nonlinearity}
\end{align}
with $\alpha\in\mathbb{R}^+$. The parameter $\alpha$ controls the slope of $w(\cdot;\alpha)$ at $r = 0.5$. A value of $\alpha > 1$ allows to increase the influence of rank errors for the loss function which will be defined next. Figure \ref{fig:transfer-function} (left) shows  $w(\cdot;\alpha)$ for different values of $\alpha$.

Using $w(\cdot;\alpha)$ and omitting $\alpha$ in the following notation we define the similarity of a sample $\boldsymbol{x_j}$ to the anchor $\boldsymbol{x_i}$ by
\begin{gather}
s_{i,j} = 1 - w(r_{i,j}) \\
s^+_{i,\max} = 1 - w(r^+_{i,\max})\quad s^-_{i,\min} = 1 - w(r^-_{i,\min})
\end{gather}
Our proposed loss function focuses on increasing $s^+_{i,\max}$ and reducing $s^-_{i,\min}$ for all anchors. The loss function is given by
\begin{equation}
J = - \frac{1}{m} \sum_{i = 1}^m \left( \log(s^+_{i,\max} + \varepsilon) + \log(1 - s^-_{i,\min} + \varepsilon) \right)
\end{equation}
Here $\varepsilon > 0$ is a small number that controls the maximum loss contribution if $s^+_{i,\max}=0$ or $s^-_{i,\min}=1$. We use $\varepsilon = 10^{-4}$ throughout the following experiments. Figure \ref{fig:transfer-function} (center) shows the influence of $\alpha$ on the loss component $-\log(s^+_{i,\max} + \varepsilon)$. 

\begin{figure}[h]
\begin{tabular*}{\linewidth}{@{\hspace{0em}}l@{\extracolsep{\fill}}r@{\hspace{0em}}}
\includegraphics[width=0.49\linewidth]{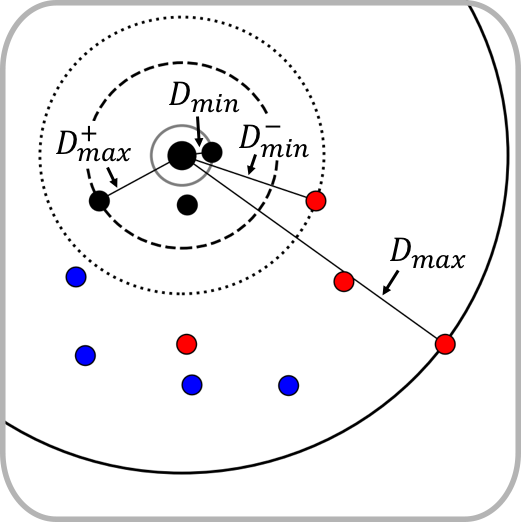} &
\includegraphics[width=0.49\linewidth]{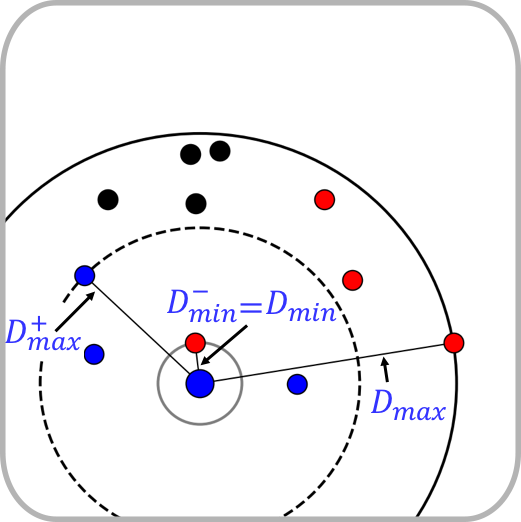}
\end{tabular*}
\centering
\caption{A toy example of a batch size $m$=$12$ (3 classes with 4 samples). Again the indices $i,j$ are omitted. As all points from the batch serve as anchors, there are 12 specific configurations, two of which are illustrated above. For each anchor point  
the four corresponding distances $D_{\min}$, $D^-_{\min}$, $D^+_{\max}$ and $D_{\max}$ are 
determined to calculate the similarities $s^+_{\max}$ and $s^-_{\min}$.
For the left black anchor point all points of the same class lie within the radius of $D^+_{\max}$ which represents a good embedding. For the right blue anchor point the closest point is from another class
leading to a high loss. The proposed loss function focuses on reducing 
$D^{+}_{\max}$ and increasing 
$D^{-}_{\min}$ for all anchors.
}
\label{fig:toy-example}
\end{figure}

Figure \ref{fig:teaser} schematically demonstrates the transformation of the original feature vector distances (a) for the proposed loss function. The anchor point is displayed as a larger red dot. The distances to the anchor are converted to approximated normalized ranks (b) using Eq. (\ref{eq:normalized-rank}). Note that the nearest point to the anchor in (a) has a rank value of 0, plotted on top of the anchor point in (b). Finally the non-linear transformation  (\ref{eq:nonlinearity}) is applied to obtain (c). Samples with higher ranks are pushed further away whereas samples with lower ranks move towards the anchor.
Figure \ref{fig:toy-example} shows two of the 12 configurations for a toy example with three classes and four samples per class.
\begin{figure*}[!hb]
    \centering
    \newlength\imgheight
    \setlength\imgheight{0.2123\textwidth}
    {\footnotesize\begin{tabular}{cccc}
    \includegraphics[height=\imgheight]{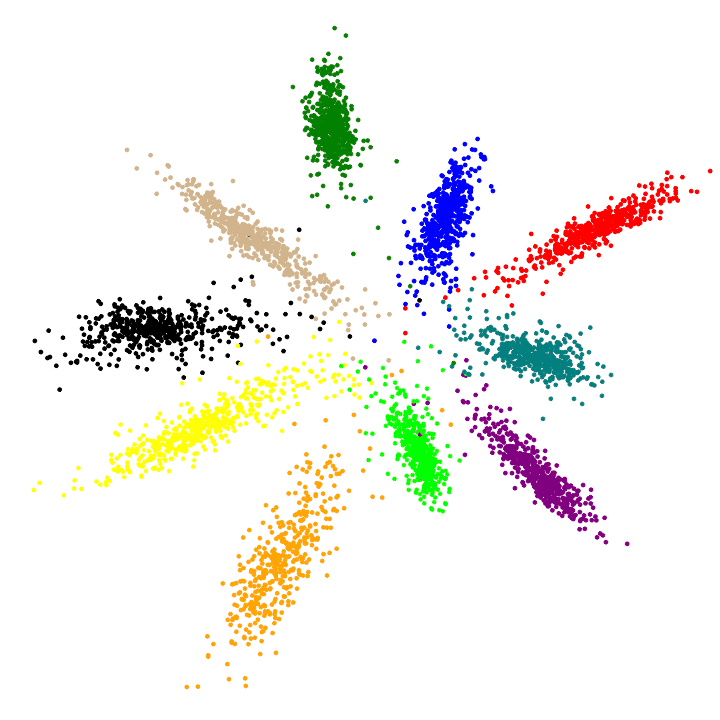} & 
    \includegraphics[height=\imgheight]{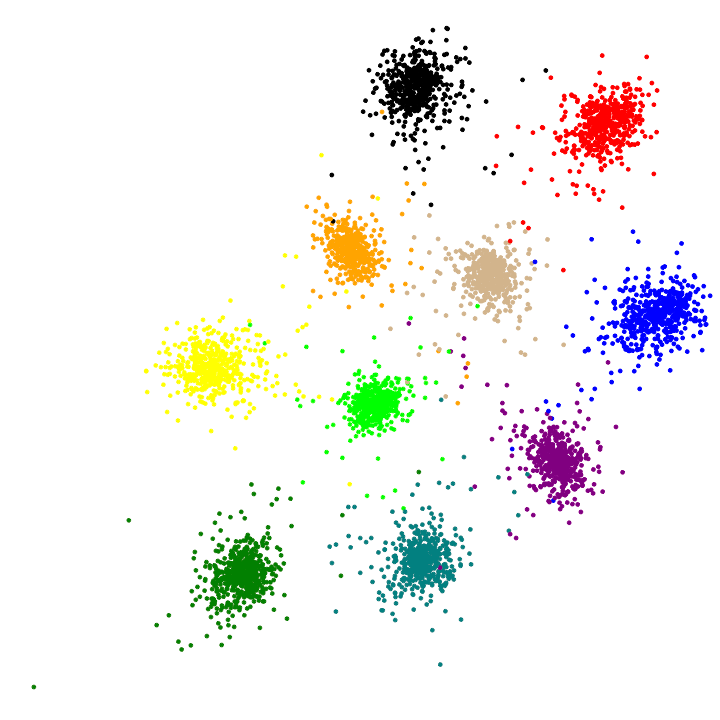} & 
    \quad \includegraphics[height=\imgheight]{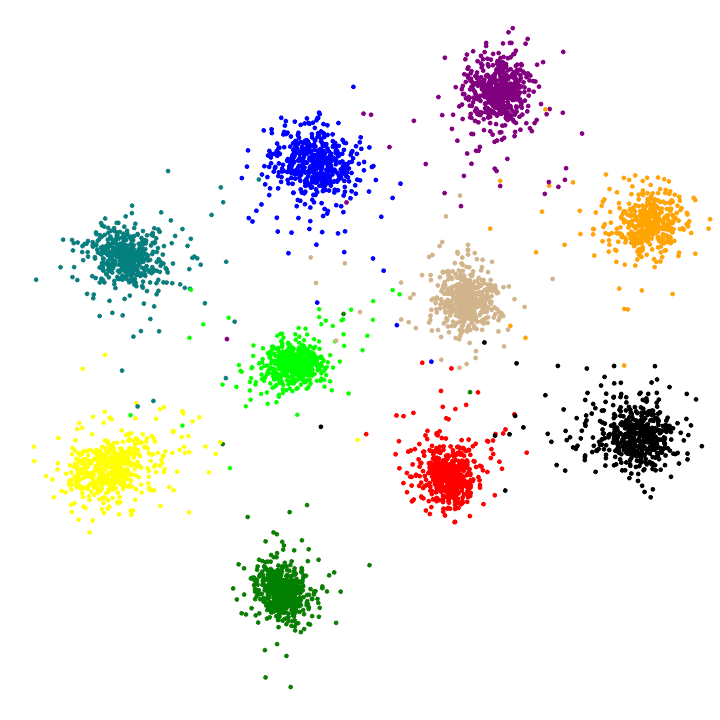} & 
    \quad \includegraphics[height=\imgheight]{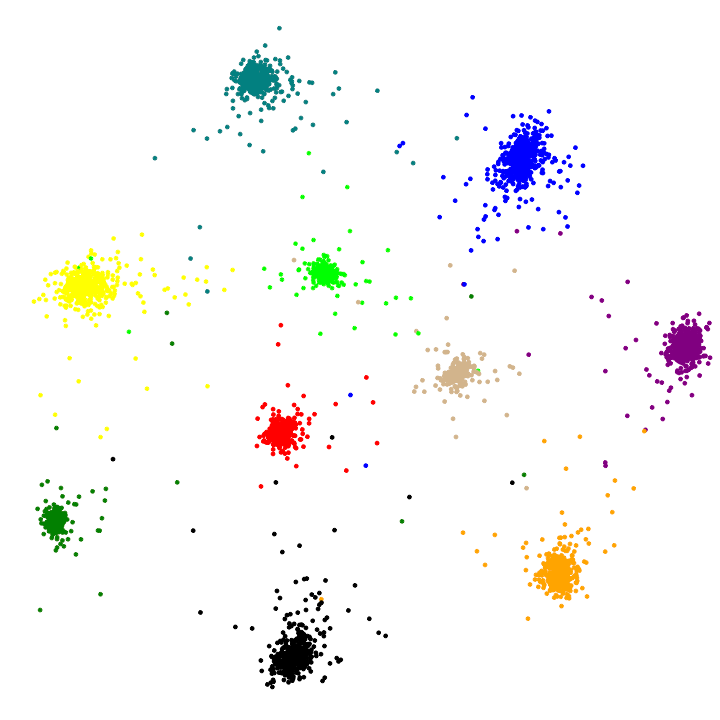} 
    \quad \includegraphics[height=\imgheight]{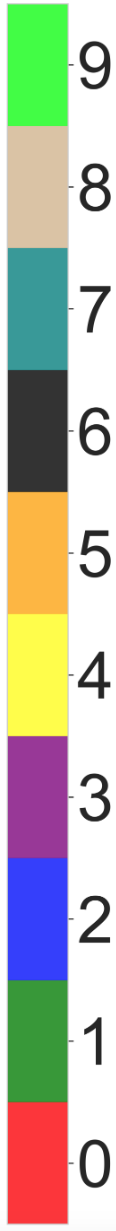} \\ 
    (a) Softmax Cross Entropy& 
    (b) Triplet Loss& 
    (c) Lifted Structured Loss& 
    (d) Nonlinear Rank Approx. Loss (ours)\\
    $\text{mAP} = 93.3$ & 
    $\text{mAP} = 97.8$ & 
    $\text{mAP} = 98.4$ & 
    $\text{mAP} = 98.8$ \\
    \end{tabular}}
    \caption{Visualization of the 2D embedding space for different loss functions using MNIST data}
    \label{fig:embedding}
\end{figure*}
\section{Experiments}
In this section, we first determine the optimal parameter $\alpha$ for our transfer function (\ref{eq:nonlinearity}) which will consequently be used throughout the other experiments. Next we compare the embeddings of the MNIST data set \cite{LeCun98}, generated by using different loss functions. Lastly, we strictly follow the evaluation protocol of \cite{Song16} and evaluate our method on three fine grained data sets, namely CUB-200-2011 \cite{CUB_200_2011}, Cars196 \cite{Car196} and Stanford Online Products \cite{Song16}. All experiments have been conducted with \textit{TensorFlow} \cite{Abadi16}, which contains implementations of the Triplet, Lifted Structured and N-pair loss functions in the contrib-package. 

\subsection{Parameterizing the Transfer Function}
The introduced family of transfer functions (\ref{eq:nonlinearity}) is parameterized by $\alpha$ as seen in Figure \ref{fig:transfer-function} (left). In the simplest case of $\alpha = 1$ the resulting function is a straight line $w(r; 1) = r$, therefore the use of $\alpha = 1$ is equivalent to optimizing our rank approximations directly. A higher $\alpha$ value leads to a lower slope in the top and bottom ranks and a steeper slope at the center. The corresponding negative log value, as used in the final loss function (\ref{eq:normalized-rank}) has a lower slope for $r \in [0, \frac{1}{2}]$ and higher values for $r \in [\frac{1}{2},1]$. The optimal value for $\alpha$ is determined empirically. For this purpose we perform a series of experiments with the Cars196 data set \cite{Car196} and $\alpha \in \{1,2,\ldots,6\}$. The exact training details are identical with the ones described in the Chapter \ref{sub:Retrieval}. As seen in Figure \ref{fig:transfer-function} (right) best results for all Recall@K values are obtained with $\alpha = 4$. We use this $\alpha$ value for all our following experiments. 

\subsection{Embedding of MNIST Data in 2D}
To demonstrate some of the appealing properties of our approach, we utilize a simple neural network with two convolutional and one fully connected layers to embed the MNIST data set \cite{LeCun98} in two dimensions. The network is first trained with the MNIST training set for exactly \numprint{20000} iterations. \numprint{5000} randomly chosen feature vectors of the test set were used for the 2D embeddings. 
Figure \ref{fig:embedding} shows the resulting embeddings achieved with the Softmax Cross Entropy, the Triplet with semi hard mining \cite{Schroff15}, the Lifted Structured and our proposed Nonlinear Rank Approximation loss functions. The Softmax Cross Entropy tries to divide the data into ten separable sections, while ignoring the actual distances. Hence it achieves a relatively low mean average precision score (mAP). The margin parameter was set to 1 for the Triplet and Lifted Structured loss functions and since both perform distance based optimization, both methods produce clusters of approximately the same size and similar intra-class distances. Even though we do not directly enforce the cluster to be dense, the representations produced by our method for every class are densely packed in the representation space. In addition, the different classes were assigned to regions of varying size, making efficient use of the available space. It appears that our method takes the original variance of each class into account and aligns the embeddings accordingly while also achieving the highest mean average precision score. 
\begin{table*}[ht]
\caption{Recall@K [\%] on CUB-200-2011, Cars196 and the Stanford Online Products data set for \numprint{64} and \numprint{512} dimensional embeddings.}
\resizebox{\textwidth}{!}{
\begin{tabular}{c||c|c|c|c||c|c|c|c||c|c|c}
    \hline
    \multirow{2}{*}{Method} & 
    \multicolumn{4}{c||}{CUB-200-2011} & 
    \multicolumn{4}{c||}{Stanford Cars196} & 
    \multicolumn{3}{c}{Stanford Online Products}\\
    \cline{2-12}
     & Recall@1 & Recall@2 & Recall@4 & Recall@8 
     & Recall@1 & Recall@2 & Recall@4 & Recall@8
     & Recall@1 & Recall@10 & Recall@100\\
    \hline\hline
    Triplet 
    & \numprint{46.3} / \numprint{51.6} & \numprint{58.6} / \numprint{63.5} & \numprint{70.9} / \numprint{74.2} & \numprint{80.2} / \numprint{83.9}
    & \numprint{56.5} / \numprint{58.4} & \numprint{69.2} / \numprint{71.4} & \numprint{78.7} / \numprint{80.2} & \numprint{86.8} / \numprint{88.0}
    & \numprint{57.2} / \numprint{59.8} & \numprint{76.1} / \numprint{78.4} & \numprint{89.4} / \numprint{91.0}
    \\
    Lifted 
    & \numprint{45.7} / \numprint{55.7} & \numprint{58.0} / \numprint{67.4} & \numprint{70.0} / \numprint{77.7} & \numprint{79.4} / \numprint{85.5} 
    & \numprint{48.8} / \numprint{50.7} & \numprint{62.3} / \numprint{63.9} & \numprint{72.8} / \numprint{73.7} & \numprint{82.3} / \numprint{83.5}
    & \numprint{61.6} / \numprint{63.8} & \numprint{79.9} / \numprint{81.4} & \numprint{91.5} / \numprint{92.7}
    \\ 
    N-Pair 
    & \numprint{51.8} / \numprint{56.4} & \numprint{63.8} / \numprint{69.1} & \numprint{74.9} / \numprint{79.1} & \numprint{83.8} / \numprint{86.6} 
    & \numprint{63.3} / \numprint{68.3} & \numprint{74.4} / \numprint{78.8} & \numprint{82.8} / \numprint{86.1} & \numprint{89.4} / \numprint{91.1}
    & \numprint{63.6} / \numprint{65.4} & \numprint{81.5} / \numprint{83.8} & \numprint{92.6} / \numprint{94.0}
    \\
    NRA (ours) 
    & \textbf{\numprint{57.6}} / \textbf{\numprint{64.3}} & \textbf{\numprint{69.4}} / \textbf{\numprint{74.8}} & \textbf{\numprint{78.8}} / \textbf{\numprint{83.4}} & \textbf{\numprint{87.3}} / \textbf{\numprint{89.9}} 
    & \textbf{\numprint{73.0}} / \textbf{\numprint{81.9}} & \textbf{\numprint{82.5}} / \textbf{\numprint{88.8}} & \textbf{\numprint{89.1}} / \textbf{\numprint{92.8}} & \textbf{\numprint{93.0}} / \textbf{\numprint{95.4}}
    & \textbf{\numprint{71.9}} / \textbf{\numprint{75.6}} & \textbf{\numprint{86.3}} / \textbf{\numprint{88.8}} & \textbf{\numprint{94.3}} / \textbf{\numprint{95.7}}
    \\ 
    \hline
\end{tabular}
}
\label{table:dataset_map}
\end{table*}
\subsection{Fine Tuning for Unseen Object Retrieval}
\label{sub:Retrieval}
In \cite{Song16} an evaluation protocol for metric learning was introduced using three fine grained data sets: CUB-200-2011 \cite{CUB_200_2011}, Cars196 \cite{Car196} and Stanford Online Products \cite{Song16}. The goal is to train with the first half of the classes and their corresponding images and to evaluate the retrieval performance with the second half.

\textbf{CUB-200-2011} contains \numprint{11788} images of \numprint{200} different bird species. The first \numprint{100} classes are chosen for training, which leads to a number of \numprint{5864} training images. The remaining \numprint{5924} samples from the second half of the classes are used for evaluation. 

\textbf{Cars196:} This collection of car images contains \numprint{196} classes and a total number of \numprint{16185} images. Again, the first half of the classes are used in the fine tuning stage and the second half, for evaluation purposes. This leads to \numprint{8054} training and \numprint{8131} test images. 

\textbf{Stanford Online Products (SOP):} The last set consists of \numprint{120053} images of \numprint{22634} different products. While the \hbox{CUB-200-2011} has an average of \numprint{58.44} images per class and the Cars196 set \numprint{82.58}, the Stanford Products data set has only \numprint{5.3} images per class. 
Again, the first half of the images is used for training and the second half for evaluation. This splits the data into \numprint{59551} training images from \numprint{11318} classes and \numprint{60502} test images from \numprint{11316} classes.

 Following the evaluation protocol from \cite{Song16} various metric learning papers have been published with incomparable results. Some utilize more advanced network architectures, while others report results for deviating embedding sizes. For this reason we have reproduced the experiments of important deep metric learning based loss functions using the GoogLeNet \cite{Szegedy15} network and feature vector sizes of \numprint{64} and \numprint{512} dimensions. The network was initialized with the parameters from the corresponding TensorFlow Slim Model \cite{TF_Slim} (\mbox{top-1} accuracy of \numprint{69.8}\% on the ImageNet validation set \cite{Russakovsky14}) and is fine tuned with the training images of each set. The network generates a set of 1024 \numprint{7}x\numprint{7} feature maps at the last convolutional layer, which are average pooled to produce a 1024 dimensional embedding. This embedding is then linearly transformed to a lower dimensional space with the help of a fully connected layer without any activation function. The input images are resized to \numprint{256} pixels at their larger dimension, before a random crop of size \numprint{224} x \numprint{224} is chosen. The crop is randomly flipped horizontally for data augmentation purposes, before it is fed into the network. The retrieval quality is measured by the Recall@K metric, as proposed in \cite{Jegou11} and is used in the evaluation protocol. This measure computes the proportion of queries in which at least one positive example is within the retrieved K nearest neighbors.

In compliance with the settings described in \cite{Song16} the number of images processed per batch is \numprint{128}. For N-pair the batches are built of \numprint{64} pairs in all three data sets. The remaining methods are trained with the \textit{uniform n-group sampling} technique. As for the CUB-200-2011 and the Cars196 data sets, \numprint{16} groups of \numprint{8} samples are chosen. Since the average number of samples per class is much lower in the Stanford Online Products data set,
\numprint{32} groups with \numprint{4} representatives are used in each batch. For a fair comparison, a stochastic gradient descent optimizer with momentum of \numprint{0.9} is used for all loss functions. A learning rate of 0.0001 was chosen for CUB-200-2011, 0.0005 for Cars196 and 0.001 for SOP. The produced feature vectors have not been $L^2$-normalized during the training phase, but they are $L^2$-normalized for testing.

Sohn \cite{Sohn16} has shown that their proposed N-pair batch sampling improves the results over the original triplet approach. The \textit{uniform n-group sampling} further improves the Triplet loss results (see Table \ref{table:dataset_map}). Since more than one positive sample per class is present, a higher number of informative triplets in the online mining process can be found. Triplet loss even outperforms Lifted Structured loss ("Lifted"), despite both using the same batching method. Our proposed Nonlinear Rank Approximation (NRA) loss function significantly outperforms the other loss functions for both dimensions and all Recall@K results.

\subsection{Comparison to the State-of-the-Art}
To put our results in context with other deep metric learning methods Table \ref{table:RecallK1} presents an overview of previously published Recall@1 values. The middle section of the table lists other not directly comparable approaches. They use either more sophisticated networks, protocol violating test/training techniques or network ensembles. Most of these modifications are applicable with our loss function and would further improve the retrieval quality. 
The superscript denotes the embedding size. In \cite{Song17} Song et al. claim the results in the N-pair \cite{Sohn16} paper have been achieved by an average of ten extracted embeddings from ten random crops. The usage of such a crop averaging technique is marked with $\boxplus$. In some cases bounding box cropping (denoted with $\square$) is applied during training and testing of the \hbox{CUB-200-2011} and Cars196 data set. Not all listed approaches employ the GoogLeNet architecture \cite{Szegedy15}. A ResNet50 v2 \cite{He16} with a top-1 accuracy of \numprint{75.6}\% on the ImageNet validation set \cite{Russakovsky14} is used by Margin and InceptionBN \cite{Ioffe15} with \numprint{73.9}\% by Proxy-NCA \cite{Movshovitz17} and Clustering \cite{Song17}. Compared to the GoogLeNet, the two more advanced architectures might give a better general image retrieval performance. If we take just the loss functions trained with GoogLeNet, using no bounding boxes and no test augmentation we can see that our Nonlinear Rank Approximation (NRA) approach achieves state-of-the-art results. It also outperforms N-pairs \cite{Sohn16} testing method and PDDM \cite{Huang16} bounding box training. In summary our method is only outperformed by attention based ensembles like ABE \cite{Kim18}. As mentioned before, HDC, A-BIER \cite{Opitz18}, \hbox{ABE-8} and Proxy-NCA \cite{Movshovitz17} use more complex/multiple network architectures. These networks are applicable in combination with our loss function and will lead to improved retrieval results. This is subject to future research.
\begin{table}[ht!]
\centering
\caption{Recall@1 [\%] results. 
The methods in the second section apply more advanced networks, testing or training techniques.}
\begin{tabular}{ c|c|c|c|c } 
 \hline
 Method & network & CUB & Cars196 & SOP \\ 
 \hline\hline
 Contrastive\textsuperscript{128} \cite{Hadsell06,Song16} & GoogLeNet & \numprint{26.4} & \numprint{21.7} & \numprint{42.0} \\
 Triplet\textsuperscript{64} \cite{Schultz03, Song16} & GoogLeNet & \numprint{36.1} & \numprint{39.1} & \numprint{42.1} \\
 Lifted Struct.\textsuperscript{128} \cite{Song16} & GoogLeNet & \numprint{47.2} & \numprint{49.0} & \numprint{60.8} \\
 Lifted Struct.\textsuperscript{512} \cite{Song16} & GoogLeNet & - & - & \numprint{62.1} \\
 Angular Loss\textsuperscript{512} \cite{Wang17} & GoogLeNet & \numprint{54.7} & \numprint{71.4} & \numprint{70.9} \\
 \hline
 Clustering\textsuperscript{64} \cite{Song17} & InceptionBN & \numprint{48.2} & \numprint{58.1} & \numprint{67.0} \\
 N-pair$\boxplus$\textsuperscript{64} \cite{Sohn16} & GoogLeNet & \numprint{51.0} & \numprint{71.1} & - \\
 N-pair$\boxplus$\textsuperscript{512} \cite{Sohn16} & GoogLeNet & - & - & \numprint{67.7} \\
 PDDM+Quad.$\square$\textsuperscript{128} \cite{Huang16} & GoogLeNet & \numprint{58.3} & \numprint{57.4} & - \\
 Proxy-NCA\textsuperscript{64} \cite{Movshovitz17} & InceptionBN & \numprint{49.2} & \numprint{73.2} & \numprint{73.7} \\
 HDC\textsuperscript{384} \cite{Yuan17} & GoogLeNet & \numprint{53.6} & \numprint{73.7} & \numprint{69.5} \\
 HTL\textsuperscript{512} \cite{Ge18} & InceptionBN & \numprint{57.1} & \numprint{81.4} & \numprint{74.8} \\
 Margin\textsuperscript{128} \cite{Manmatha17} & ResNet50v2 & \numprint{63.6} & \numprint{79.6} & \numprint{72.7} \\
 A-BIER\textsuperscript{512} \cite{Opitz18} & GoogLeNet & \numprint{57.5} & \numprint{82.0} & \numprint{74.2} \\
 ABE-8\textsuperscript{512} \cite{Kim18} & GoogLeNet & \numprint{60.6} & \numprint{85.2} & \numprint{76.3} \\
 \hline
 NRA (ours) \textsuperscript{64} & GoogLeNet & \numprint{57.6} & \numprint{73.0} & \numprint{71.9} \\ 
 NRA (ours) \textsuperscript{512} & GoogLeNet & \numprint{64.3} & \numprint{82.1} & \numprint{75.6} \\ 
 \hline
\end{tabular}
\label{table:RecallK1}
\end{table}

\section{Conclusion}
In this paper, we propose Nonlinear Rank Approximation loss (NRA) for deep metric learning, which significantly improves upon existing approaches like Triplet, Lifted Structured, and N-pair loss. Our approach uses all batch elements as anchor points. Instead of having to deal with all other batch elements for each anchor we focus on those two that mostly impact the clustering and retrieval performance: the furthest point of the same class and the closest point of a different class as the anchor. By using nonlinear transformed ranks instead of distances our proposed loss function is better able to optimize the embedding for image retrieval tasks. We demonstrate the effectiveness of the new approach on fine-grained visual recognition, as well as visual object clustering and retrieval. The NRA loss function generates very compact embeddings, therefore this new approach can also be used for dimensionality reduction and feature compression.

\bibliographystyle{abbrv}
\bibliography{mmsp2019}

\end{document}